\documentclass{article}
\PassOptionsToPackage{numbers, sort&compress}{natbib}

\usepackage[main, final]{neurips_2026}

\usepackage[utf8]{inputenc} 
\usepackage[T1]{fontenc}    
\usepackage{hyperref}       
\usepackage{url}            

\usepackage{booktabs}       
\usepackage{amsfonts}       
\usepackage{nicefrac}       
\usepackage{microtype}      
\usepackage{xcolor}         
\usepackage{wrapfig}
\usepackage{makecell}
\usepackage{graphicx}
\usepackage{caption}
\usepackage{multirow}

\usepackage{longtable}
\usepackage{tabularx}
\usepackage{amsmath}
\usepackage{hhline}
\title{RotVLA: Rotational Latent Action for Vision-Language-Action Model}

\author{%
  Qiwei Li$^{1,2}$
  \quad
  Xicheng Gong$^{1}$
  \quad
  Xinghang Li$^{2}$
  \quad
  Peiyan Li$^{3}$
  \\
  \textbf{Quanyun Zhou}$^{2}$
  \quad
  \textbf{Hangjun Ye}$^{2}$
  \quad
  \textbf{Jiahuan Zhou}$^{1*}$
  \quad
  \textbf{Yadong Mu}$^{1}$\thanks{Corresponding Author}
  \\
  \\
  $^1$Wangxuan Institute of Computer Technology, Peking University 
  \\
  $^2$Xiaomi Robotics
  \quad
  $^3$CASIA
  \\
  \\
  \url{https://qiweili00.github.io/rotvla-page/}
}

\begin{document}

\maketitle

\begin{abstract}
  Latent Action Models (LAMs) have emerged as an effective paradigm for handling heterogeneous datasets during Vision-Language-Action (VLA) model pretraining, offering a unified action space across embodiments. However, existing LAMs often rely on discrete quantization encode and decode pipelines, which can lead to trivial frame reconstruction behavior, limited representational capacity, and a lack of physically meaningful structure. We introduce \textbf{RotVLA}, a VLA framework built on a continuous rotational latent action representation. Latent actions are modeled as elements of ${\rm SO}(n)$, providing continuity, compositionality, and structured geometry aligned with real-world action dynamics. A triplet frame learning framework further enforces meaningful temporal dynamics while avoiding degeneration. RotVLA consists of a VLM backbone and a flow-matching action head, pretrained on large-scale cross-embodiment robotic datasets and human videos with latent-action supervision. For downstream robot control, the flow-matching head is extended into a unified action expert that jointly denoises latent and robot actions. Here, latent actions serve as a latent planner, providing high-level guidance that conditions action generation. With only 1.7B parameters and 1700+ hours of pretraining data, RotVLA achieves 98.2\% on LIBERO and 89.6\% / 88.5\% on RoboTwin2.0 under clean and randomized settings, respectively. It also demonstrates strong real-world performance on manipulation tasks, consistently outperforming existing VLA models.
\end{abstract}

\section{Introduction}
\label{sec:intro}

\begin{figure}[t]
  \centering
  \includegraphics[height=5.0cm]{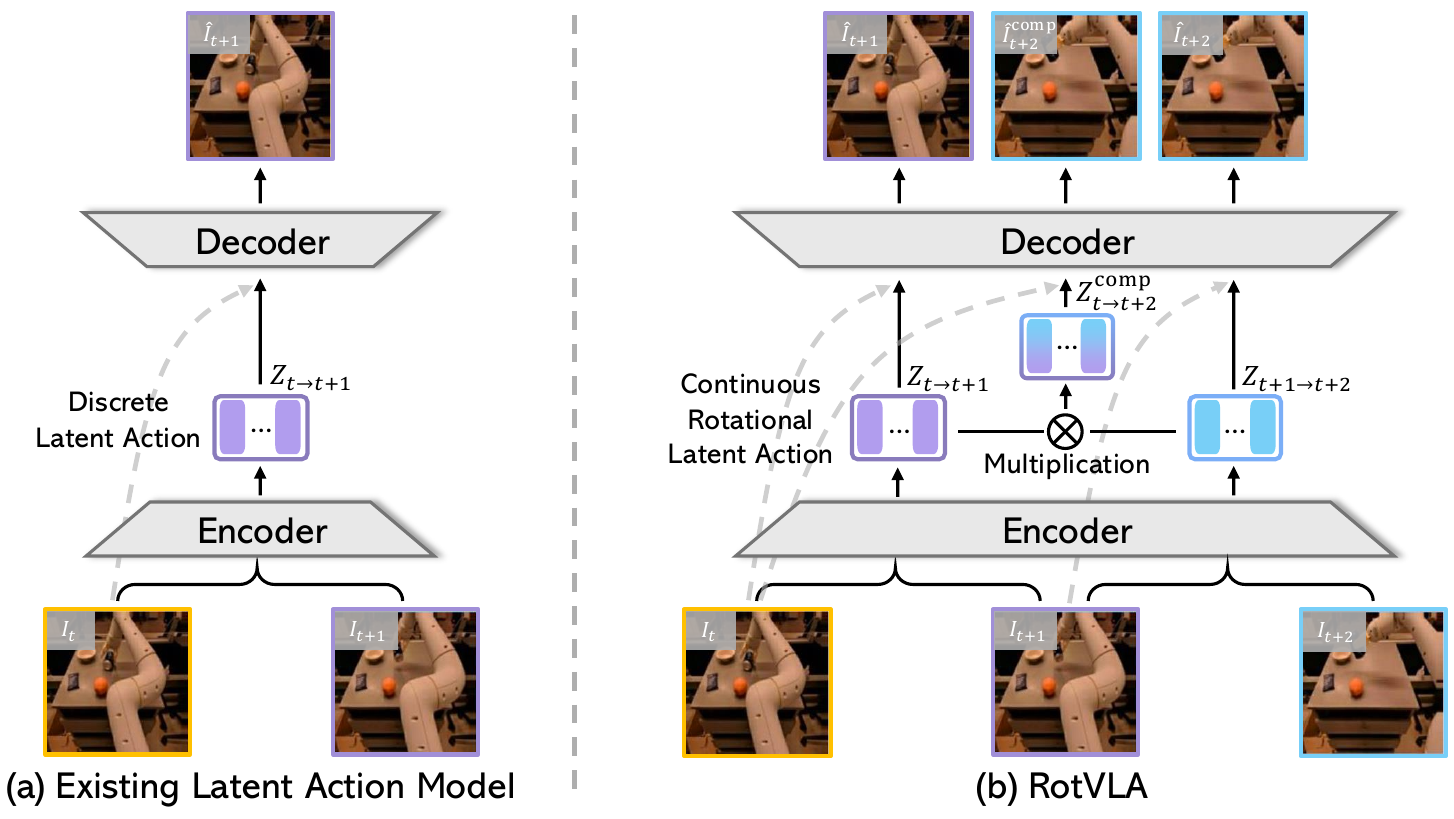}
  \caption{Illustration of existing LAMs (a) and RotVLA (b). Existing LAMs follow an encode–decode paradigm and model latent actions as discrete tokens. In contrast, RotVLA represents latent actions as continuous elements of ${\rm SO}(n)$. Additionally, we introduce a triplet learning framework that not only learns latent actions between consecutive frames but also composes these actions to predict future frames, reinforcing the capture of meaningful temporal dynamics.
  }
  \label{fig:introduction}
\end{figure}

Recently, Vision-Language-Action (VLA)~\cite{ma2024survey,sapkota2025vision} models that are pretrained from Vision-Language Models (VLMs)~\cite{liu2023visual,chen2024internvl,bai2025qwen2} using large-scale robot-related data have demonstrated substantial potential. However, VLA pretraining faces a fundamental challenge: effectively handling heterogeneous data collected from diverse robot embodiments and unlabeled human videos. A promising direction to address these issues is Latent Action Models (LAMs)~\cite{ye2024latent,bu2025univla,bu2025agibot}, which aim to learn a unified action space shared across datasets. Most existing LAMs adopt an encode and decode paradigm (Fig.\ref{fig:introduction}(a)), and typically apply quantization methods such as VQ-VAE~\cite{van2017neural} to the latent action space. This constrains the space and makes it compatible with the next-token prediction training paradigm of large VLMs. However, this paradigm suffers from three key limitations: (1) LAMs have the risk of degenerating into trivial solutions that simply encode the target frame~\cite{liang2025clam}. (2) Discretizing the latent action space severely restricts its representational capacity and contradicts the inherently continuous nature of physical actions. (3) The latent action space lacks physical meaning, providing no notion of scale or meaningful action composition.

\begin{figure}[t]
  \centering
  \includegraphics[width=1\linewidth]{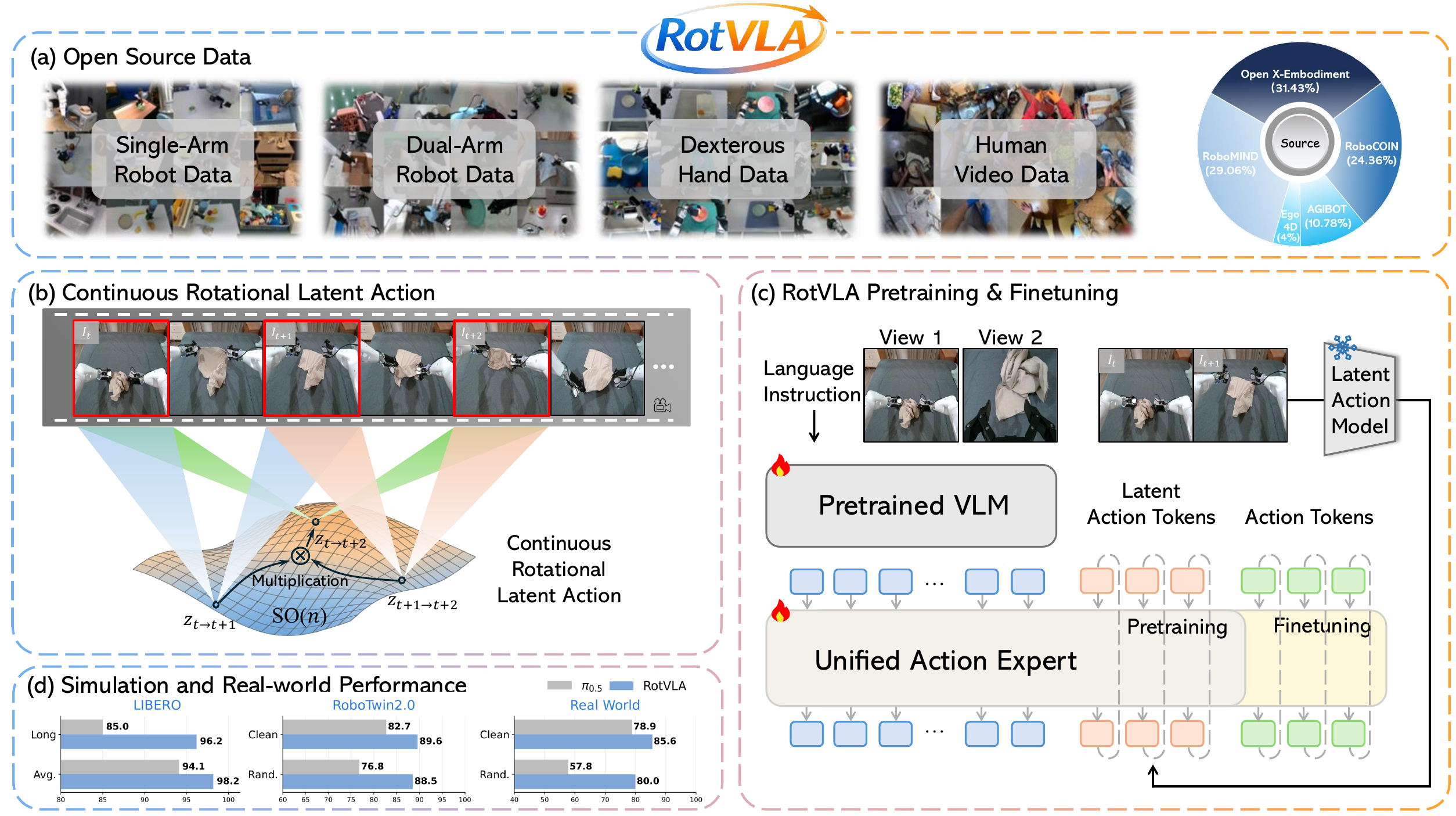}
  \caption{We introduce RotVLA, a Vision-Language-Action framework pretrained with a continuous rotational latent action on over 1700 hours of cross-embodiment robot and human video data. The latent actions are trained via triplet learning framework and further integrated into a unified flow-matching action expert, where they serve as high-level planners guiding robot control. Results demonstrate strong performance across simulation benchmarks and real-world manipulation tasks.}
  \label{fig:pipeline}
  \vspace{-10pt}
\end{figure}

To address these issues, we propose \textbf{RotVLA}, a VLA framework built upon a continuous rotational latent action representation. As shown in Fig.\ref{fig:pipeline}, instead of modeling latent actions as discrete tokens, we represent them as elements of the $n$-dimensional rotation group, ${\rm SO}(n)$. This formulation naturally aligns with the continuous nature of robot actions and provides substantially richer representational capacity compared to discrete tokens. Moreover, due to the closure property of ${\rm SO}(n)$, latent-action composition can be defined via matrix multiplication, mirroring real-world action composition.

Leveraging this representation, we formulate latent action learning as transitions across three frames (Fig.~\ref{fig:introduction}(b)). Given a frame triplet, we apply standard encoding and decoding to each pair of consecutive frames and adopt a soft quantization mechanism~\cite{chen2025softvq} that constrains the latent space while preserving continuity. We then compose the latent actions from the first and second transitions to infer the latent action from the first to the third frame, and supervise the prediction. This objective encourages the encoder to capture frame-to-frame dynamics rather than relying on direct reconstruction shortcuts.

After pretraining the latent action model, any pair of consecutive video frames can be mapped into a continuous ${\rm SO}(n)$ latent action. Building on this, we construct RotVLA, which integrates a pretrained VLM backbone with a flow-matching latent action head. RotVLA is pretrained to predict latent actions using large-scale human and cross-embodiment robot datasets.

To adapt RotVLA to downstream manipulation tasks, we extend the latent action head into a unified flow-matching action head that simultaneously denoises latent actions and robot actions. Through causal attention, robot-action generation is conditioned on latent actions, allowing the latent space to function as the planner that guides action prediction.

Our key contributions are summarized as follows:

(1) We propose a continuous rotational formulation of latent action based on ${\rm SO}(n)$, which captures the continuous and compositional properties of real-world actions. The latent action is learned from the dynamics across three consecutive frames and provides a unified action representation that generalizes across embodiments with zero-shot action generation capabilities.

(2) We introduce RotVLA, a VLA pretrained on large-scale human and cross-embodiment robot datasets using a flow-matching latent action expert. For fine-tuning on downstream tasks, we design a unified action expert that treats latent action as a planner to guide robot action prediction.

(3) With only 1.7B parameters, RotVLA achieves 98.2\% on LIBERO~\cite{liu2023libero} and 89.6\% / 88.5\% on RoboTwin2.0~\cite{chen2025robotwin} benchmark. Real-world experiments on tasks such as pick-and-place and cup stacking demonstrate improvements over existing VLA methods.

\section{Related Work}
\subsection{Vision-Language-Action Model}

With the rapid progress of Vision-Language Models (VLMs)~\cite{liu2023visual,chen2024internvl,bai2025qwen2}, multimodal systems have demonstrated strong capabilities in perception and semantic understanding across both images and text. VLMs provide a unified embedding space that aligns visual observations with linguistic concepts, enabling powerful generalization, instruction following, and cross-modal reasoning. Building upon this foundation, Vision-Language-Action (VLA) models~\cite{ma2024survey,sapkota2025vision,li2024towards} extend VLMs into the action domain by integrating control within the multimodal framework.

Typically, VLA systems adopt a two-stage paradigm in which a pretrained VLM is first adapted to robot-specific multimodal data, followed by fine-tuning on target environments or tasks. Some prior works~\cite{kim2024openvla,li2025bridgevla,pertsch2025fast,bu2025univla} follow a next-token prediction formulation, where continuous actions are discretized into bins or encoded as action tokens, allowing pretraining and fine-tuning using standard language modeling techniques. To better preserve the continuous nature of actions, other approaches~\cite{wang2025vla,kim2025fine} directly model actions as continuous values and train the model using regression objectives, achieving strong performance during fine-tuning. Beyond token-based and regression-based formulations, a growing body of research~\cite{intelligence2025pi,intelligence2504pi0,bu2025agibot,bjorck2025gr00t,lin2025evo,liu2024rdt,bi2025hrdt,zheng2025x} explores generative modeling techniques for action prediction. These methods employ diffusion-based or flow-matching policies that treat action generation as a conditional generative process, producing smooth and coherent action sequences.

\subsection{Latent Action Model}

Most VLA methods rely on large-scale, robot-specific annotated datasets for pretraining. However, learning from cross-embodiment datasets remains challenging due to substantial variations in visual observations, action spaces, and embodiment configurations~\cite{zheng2025x}. Furthermore, these approaches rely on ground-truth action annotations, which limits their ability to leverage large-scale Internet video data. To address these issues, a line of work focuses on Latent Action Models, which aim to learn a unified action space across heterogeneous datasets by encoding the transition between sequential observations. Mainstream approaches~\cite{ye2024latent,bruce2024genie,chen2024igor,chen2024moto,li2025latbot,yang2025como} typically adopt an Inverse Dynamics Model (IDM) to infer latent actions from consecutive video frames, coupled with a Forward Dynamics Model (FDM) that reconstructs future observations conditioned on the inferred latent action. Building on this formulation, \cite{chen2025villa,zhang2026clap} incorporate annotated actions to guide latent-action learning, while \cite{bi2025motus,bjorck2025gr00t} treat latent actions as surrogate labels for unlabeled data. Other works enhance these models by introducing additional modalities such as depth or optical flow~\cite{cai2025seeing,bi2025motus,bu2025laof,govind2026unilact}.

Despite their promise, these pipelines have the risk of degenerating into trivial solutions that simply encode and reconstruct the future frame~\cite{liang2025clam}. Existing methods attempt to mitigate this issue by limiting the latent action space by quantization, thereby encouraging the model to capture the underlying dynamics between frames. However, such discretization contradicts the inherently continuous and compositional nature of actions. To address this, we introduce a continuous latent action representation based on elements of ${\rm SO}(n)$, together with a triplet learning framework that explicitly enforces the latent action to model the underlying dynamics between frames.

\section{RotVLA}

\subsection{Stage \text{I}: Continuous Rotational Latent Action}
\textbf{Preliminary.} A typical LAM consists of an inverse dynamics encoder $\mathcal{E}$ and a forward dynamics decoder $\mathcal{D}$. Given two consecutive video frames with $k$ interval, denoted as $I_t$ and $I_{t+1}$, the encoder extracts a latent action representation composed of $N$ tokens of dimension $d$, i.e. $z_{t\rightarrow t+1} \in \mathbb{R}^{N \times d}$. In conventional LAMs, this latent action is obtained through vector quantization methods VQ-VAE~\cite{van2017neural}:
\begin{equation}
    z_{t\rightarrow t+1} = VQ\big(\mathcal{E}(I_t,I_{t+1})\big).
    \label{eq: latent action encode raw}
\end{equation}
The decoder then predicts the next frame conditioned on the latent action and the current frame:
\begin{equation}
    \hat{I}_{t+1} = \mathcal{D}(I_t,z_{t\rightarrow t+1}).
\end{equation}
The LAM is supervised by minimizing the reconstruction error:
\begin{equation}
    \mathcal{L}_{{\rm LAM}} = || \hat{I}_{t+1} - I_{t+1} ||_2.
\end{equation}

\textbf{Continuous latent action.} As discussed earlier, existing LAMs suffer from discontinuities introduced by hard quantization, which limits their ability to model inherently continuous physical actions. To address this issue, we replace VQ-VAE~\cite{van2017neural} with SoftVQ~\cite{chen2025softvq}, which aggregates multiple codewords through a soft categorical distribution. This mechanism preserves continuity in the latent space while retaining codebook structure. Details about SoftVQ are provided in the Appendix.

\textbf{Rotational latent action.} Building upon the continuous formulation, we further constrain the latent action to lie in the rotation group ${\rm SO}(n)$, yielding an ${\rm SO}(n)$-valued latent representation. Specifically, we represent each latent action as an $n \times n$ rotation matrix. We empirically find that directly enforcing the latent action to match the form of a rotation matrix overly constrains its representational capacity. Therefore, the model first predicts an unconstrained matrix, $M = U\Sigma V^{\top}$. We then obtain the closest rotation matrix in Frobenius norm through SVD-based orthogonal projection, defined as ${\rm Proj}(\cdot)$:
\begin{equation}
    {\rm Proj}(M) = U {\rm diag}\big( 1,1,...,{\rm det}(UV^{\top})\big)V^{\top}.
\end{equation}
This projection ensures that the latent action satisfies orthogonality and unit determinant constraints while maintaining expressive flexibility during learning. For notational simplicity, we define the latent action extraction operator:
\begin{equation}
    \mathcal{F}(I_t,I_{t+1}) =  {\rm Proj}\Big(VQ_{\rm soft}\big(\mathcal{E}(I_t,I_{t+1})\big)\Big),
\end{equation}
where $VQ_{\rm soft}$ represents the SoftVQ quantization. Thus, the LAM learning with continuous rotational latent action with frame $I_t,I_{t+1}$ in Eq.\ref{eq: latent action encode raw} can be rewritten as:$z_{t \rightarrow t+1} = \mathcal{F}(I_t,I_{t+1}).$

\textbf{Triplet learning framework.} To prevent degeneration into trivial frame reconstruction solutions, we introduce a triplet learning framework for latent action modeling. Given three consecutive frames sampled with interval $k$, denoted as $I_t,I_{t+1}$ and $I_{t+2}$, the framework extends the standard two-frame LAM objective by explicitly enforcing temporal compositionality across two transitions. We first extract the corresponding latent actions using the encoder: 
\begin{equation}
    z_{t\rightarrow t+1} = \mathcal{F}(I_t, I_{t+1}), \quad
z_{t+1\rightarrow t+2} = \mathcal{F}(I_{t+1}, I_{t+2}).
\label{eq: latent action}
\end{equation}
Following standard LAM training, single-step predictions are supervised via reconstruction losses:
\begin{equation}
    \hat{I}_{t+1} = \mathcal{D}(I_t, z_{t\rightarrow t+1}), \quad
    \hat{I}_{t+2} = \mathcal{D}(I_{t+1}, z_{t+1\rightarrow t+2}),
\end{equation}
\vspace{-10pt}
\begin{equation}
    \mathcal{L}_{\rm single} = \|\hat{I}_{t+1} - I_{t+1}\|_2^2 + \|\hat{I}_{t+2} - I_{t+2}\|_2^2.
\end{equation}

The latent action space admits a gauge ambiguity: without additional constraints, the identity transformation is not guaranteed to coincide with the canonical identity element of SO(n). We therefore explicitly anchor the identity latent action using stable frame pairs. Concretely, we compute the batch-wise mean latent action from identical frame pairs and project it onto ${\rm SO}(n) $ to obtain an identity element $z_{\mathcal{I}}$:
\begin{equation}
    z_{\mathcal{I}} = {\rm Proj} \Big( \mathbb{E} \big( \mathcal{F}(I_t, I_t) \big) \Big).
\end{equation}

Since each action lies on ${\rm SO}(n)$, action composition naturally corresponds to matrix multiplication. The composed two-step latent action is defined as:
\begin{equation}
    z_{t\rightarrow t+2}^{\rm comp} = z_{t+1\rightarrow t+2} \cdot z_{\mathcal{I}}^{-1} \cdot z_{t\rightarrow t+1}.
\end{equation}
We then generate a two-step prediction by applying the composed action to the first frame:
\begin{equation}
    \hat{I}_{t+2}^{\rm comp} = \mathcal{D}(I_t,z_{t\rightarrow t+2}^{\rm comp}).
\end{equation}
A reconstruction loss penalizes the difference between this prediction and the ground-truth frame $I_{t+2}$ as $\mathcal{L}_{\rm comp} = \|\hat{I}_{t+2}^{\rm comp} - I_{t+2}\|_2^2.$ The overall training loss:
\begin{equation}
    \mathcal{L}_{\rm triplet} = \mathcal{L}_{\rm single} + \mathcal{L}_{\rm comp}+\mathcal{L}_{\rm soft},
\end{equation}
where $\mathcal{L}_{\rm soft}$ is the Kullback-Leibler loss in SoftVQ~\cite{chen2025softvq} for codebook learning.

The objective $\mathcal{L}_{\rm triplet}$ not only promotes accurate single step frame reconstruction through $\mathcal{L}_{\rm single}$, but also enforces structural consistency under action composition via $\mathcal{L}_{\rm comp}$. By enforcing that the composition of $z_{t\rightarrow t+1}$ and $z_{t+1\rightarrow t+2}$ reproduce $I_{t+2}$, the model is explicitly penalized if the latent actions degenerate to trivial solutions that only reconstruct the immediate next frame. In this way, the triplet learning framework promotes the learning of meaningful dynamic representations that capture temporal transitions.



\subsection{Stage \text{II}: RotVLA Pretraining}
\label{sec: rotvla pretraining}
After learning the continuous rotational latent action space, the latent actions are used as supervision for training the RotVLA. Following prior VLA frameworks~\cite{bjorck2025gr00t,zheng2025x}, we construct RotVLA by combining a pretrained VLM backbone with a diffusion transformer (DiT) based action expert. We adopt InternVL3.5-1B~\cite{wang2025internvl35} as the VLM model. The overall architecture of RotVLA is illustrated in Fig.\ref{fig:pipeline}. Given a pair of consecutive frames $(I_t,I_{t+1})$ along with language instruction, the VLM processes this multimodal input and produces the corresponding representation of vision and language tokens, $h$. Meanwhile, the corresponding latent action tokens $z_{t\rightarrow t+1} \in \mathbb{R}^{n \times n}$ are obtained via Eq.\ref{eq: latent action}. These latent actions are implemented as an $n$-dimensional action chunk with horizon $n$. RotVLA is then pretrained using a flow-matching objective~\cite{zheng2025x,intelligence2504pi0}. Specifically, the action expert predicts a velocity field, $v_\theta(z_\tau,\tau,h)$, conditioned on the VLM features $h$, where $z_\tau = \tau z_{t \rightarrow t+1} + (1-\tau) z_0,z_0 \sim \mathcal{N}(0,{\rm I}),\tau \in [0,1]$ is the time index. Flow-matching minimizes the squared error between predicted and true velocities:

\begin{equation}
    \mathcal{L}_{\rm LA}^{\rm FM}= \mathbb{E}_{\tau, z_{t \rightarrow t+1}, z_0}\left[\left\|
            v_\theta(z_\tau, \tau, h) - (z_{t \rightarrow t+1} - z_0)
        \right\|_2^2
    \right].
\end{equation}

\subsection{Stage \text{III}: RotVLA Finetuning}
After the pretraining stage, RotVLA has been empowered with embodiment knowledge, by learning a multimodal model that maps actions from diverse robotic embodiments into a unified latent-action space. To adapt the pretrained model to real robotic manipulation, we introduce a unified flow-matching action expert that jointly predicts: (1) latent actions $z \in {\rm SO}(n)$, serving as a latent planner; (2) $d$-dimensional robot action chunk with horizon $N$, $a\in \mathbb{R}^{N \times d}$, serving as robot controller. Following the procedure described in Sec.\ref{sec: rotvla pretraining}, we extract latent actions and define a joint action variable, $x=(a,z_{t\rightarrow t+1})$. We then apply a unified flow-matching objective to jointly supervise latent and robot actions:
\begin{equation}
    \mathcal{L}_{\rm LA-RA}^{\rm FM}= \mathbb{E}_{\tau, x, x_0}\left[\left\|
            v_\theta(x_\tau, \tau, h) - (x - x_0)
        \right\|_2^2
    \right],
\end{equation}
where $x_\tau = \tau x + (1-\tau) x_0,x_0 \sim \mathcal{N}(0,{\rm I}),\tau \in [0,1]$ is the time index. This formulation enables simultaneous denoising of latent planning variables and embodiment-specific control actions within a unified diffusion framework.

\textbf{Structured attention mechanism.} To ensure proper information flow between perception, latent planning, and robot control, we introduce a structured attention mechanism within the action head. Specifically, latent action tokens attend only to vision-language tokens, without attending to real action tokens aligning with the pretraining stage. While the robot action tokens attend to both latent action tokens and vision–language tokens. This structured attention enforces a clear separation between planning and control, while maintaining effective cross-modal interaction. By preserving latent action during finetuning, RotVLA retains the action semantics learned from cross-embodiment data, while enabling efficient adaptation to new robotic platforms and downstream tasks.

\subsection{Data}
To construct the pretraining datasets for both LAM and RotVLA, we aggregate a large-scale dataset, including Open X-Embodiment~\cite{o2024open}, AGIBOT-beta~\cite{bu2025agibot}, RoboMIND~\cite{wu2024robomind}, RoboCOIN~\cite{wu2025robocoin} and Ego4D~\cite{grauman2022ego4d}, spanning diverse single-arm, bi-manual robots, dexterous hands and human video data over 1700 hours. Detailed dataset statistics are provided in the Appendix.

\begin{table}[t]
\caption{Evaluation on simulation benchmarks (LIBERO and RoboTwin2.0). * denotes models pretrained with latent actions, while $\dagger$ denotes models pretrained with both latent and robot actions.}
\label{table: libero results}
\centering
\small

\renewcommand{\arraystretch}{1.1}
\begin{tabularx}{0.96\textwidth}{l>{\centering\arraybackslash}m{0.88cm}|>{\centering\arraybackslash}m{0.88cm}>{\centering\arraybackslash}m{0.88cm}>{\centering\arraybackslash}m{0.88cm}>{\centering\arraybackslash}m{0.88cm}>{\centering\arraybackslash}m{0.88cm}|>{\centering\arraybackslash}m{0.88cm}>{\centering\arraybackslash}m{0.88cm} }
\hline
\multirow{2}{*}{\textbf{Methods}} & \multirow{2}{*}{\textbf{Size}}
& \multicolumn{5}{c|}{\textbf{LIBERO}} 
& \multicolumn{2}{c}{\textbf{RoboTwin2.0}}  \\
 &  
 & Spatial & Object & Goal & Long & Avg. 
 & Clean & Rand. \\
\hline



GO-1$^\dagger$~\cite{bu2025agibot} &  2B & 96.2  &  97.8 &  96.0 &  89.2 & 94.8    & 37.8 & 36.2\\
CoMo*~\cite{yang2025como} & 1B &  80.3 & 95.0 & 85.0 & 63.0 & 80.8    & - & -   \\
OpenVLA~\cite{kim2024openvla}  & 7B &  84.7 &88.4 &79.2 &53.7 &76.5 & - & -  \\
OpenVLA-OFT~\cite{kim2025fine}  &  7B & 97.6 &98.4 &97.9 &94.5 &97.1 & - & -  \\
$\pi_0$~\cite{black2024pi_0} & 3B & 96.8 & 98.8 & 95.8 & 85.2 & 94.2 & - & -  \\
$\pi_{0.5}$~\cite{intelligence2504pi0} & 3B & 96.8 & 98.8 & 95.8 & 85.2 & 94.1 &82.7 &  76.8 \\
GR00T-N1.6$^\dagger$~\cite{bjorck2025gr00t} & 3B & 97.7 & 98.5 & 97.5 &94.4 & 97.0 & - & -  \\

UniVLA*~\cite{bu2025univla} & 9B &  95.4 & 98.8 & 93.6 & 94.0 & 95.4 & - & -   \\
villa-X$^\dagger$~\cite{chen2025villa} & 3B &  97.5 & 97.0 & 91.5 & 74.5 & 90.1 & - & -   \\

X-VLA~\cite{zheng2025x} & 0.9B & \textbf{98.2} & 98.6 & 97.6 & \textbf{97.8} & 98.1 & 72.8 & 72.8\\
StarVLA~\cite{community2026starvla} & 4B & 97.8 & 98.6 & 96.2 & 93.8 & 96.6 & 88.2 & 88.3 \\
Motus$^\dagger$~\cite{bi2025motus} & 8B & - & - & - & \textbf{97.8} & - & 88.7 & 87.0\\
LingBot-VLA~\cite{wu2026pragmatic} & 4B & - & - & - & - & - & 88.6 & 86.7\\

\hline 
\textbf{RotVLA} & 1.7B & \textbf{98.2} & \textbf{99.6} & \textbf{98.4} & 96.4 & \textbf{98.2} & \textbf{89.6} & \textbf{88.5}\\

\hline
\end{tabularx}
\end{table}

\section{Experiment}

\subsection{Implementation Details}
\label{sec: Implementation Details}
\textbf{Model architecture.} Following UniVLA~\cite{bu2025univla}, we implement the LAM encoder using a frozen DINOv2~\cite{oquab2023dinov2} to extract frame-wise visual features. These features are further processed by a spatial–temporal transformer~\cite{xu2020spatial}. The decoder is implemented with standard transformer~\cite{dosovitskiy2020image}. The latent action dimension $n$ is set to 16. For RotVLA, the VLM backbone is initialized from InternVL3.5-1B~\cite{wang2025internvl35}, followed by an action expert implemented as a 24-layer Diffusion Transformer (DiT)~\cite{peebles2023scalable}. The full model contains approximately 1.7B, including 304M in the vision encoder, 752M in the language model, 305M in the action head, and 290M in the latent action model.

\textbf{Training details.}
We pretrain both LAM and RotVLA for 200k steps, with a batch size of 256. During downstream finetuning, RotVLA is trained with a batch size of 128. For robot action representation during finetuning, we adopt the absolute end-effector (abs EEF) pose formulation, consisting of xyz position and Rotate6D~\cite{zhou2019continuity} representation for orientation, which aligns naturally with our rotational latent action formulation. The model is optimized using the AdamW optimizer with a learning rate of 1e-4 and weight decay of 0.01. During RotVLA pretraining and finetuning, we adopt the learning rate strategy proposed in X-VLA~\cite{zheng2025x}: the VLM backbone is assigned a reduced learning rate to mitigate catastrophic forgetting, and the action expert is warmed up for the first 5k steps while keeping the VLM frozen. Pretraining is conducted on 8 NVIDIA H200 GPUs for 50 hours. Additional implementation details are provided in the Appendix.
\subsection{Simulation Results}
\label{sec:Simulation Results}
We evaluate RotVLA on two widely-used benchmarks:
LIBERO~\cite{liu2023libero} and RoboTwin2.0~\cite{chen2025robotwin}.

\textbf{LIBERO benchmark.} The LIBERO benchmark~\cite{liu2023libero} is a multi-task robotic manipulation suite designed to evaluate manipulation ability across diverse tasks and scene configurations. It includes four suites: LIBERO-Spatial, LIBERO-Object, LIBERO-Goal and LIBERO-Long, each consists of 10 tasks. Following OpenVLA~\cite{kim2024openvla}, we filter out unsuccessful trajectories from the training data. RotVLA is finetuned jointly on the four suites for 80k steps, yielding a single multi-task model capable of handling all tasks. As shown in Table.\ref{table: libero results}, RotVLA achieves an average success rate of \textbf{98.2\%}, surpassing other strong baselines such as UniVLA, GR00T-N1.6, and X-VLA. 

\textbf{RoboTwin2.0 benchmark.} RoboTwin2.0~\cite{chen2025robotwin} is a large-scale simulation benchmark for evaluating bimanual robotic manipulation. The benchmark instantiates a suite of 50 dual-arm collaboration tasks. Following the evaluation protocol of ~\cite{bi2025motus,wu2026pragmatic}, we train a single model across all 50 tasks. Each task includes 50 clean demonstrations and 500 randomized demonstrations. During evaluation, we execute 100 rollouts per task under different random seeds. As reported in Table.\ref{table: libero results}, RotVLA achieves a success rate of \textbf{89.6\%} and \textbf{88.5\%} on the clean and randomized setting, outperforming prior VLA baselines. Detailed results are provided in Appendix.

\begin{figure}[t]
  \centering
  \includegraphics[width=0.95\linewidth]{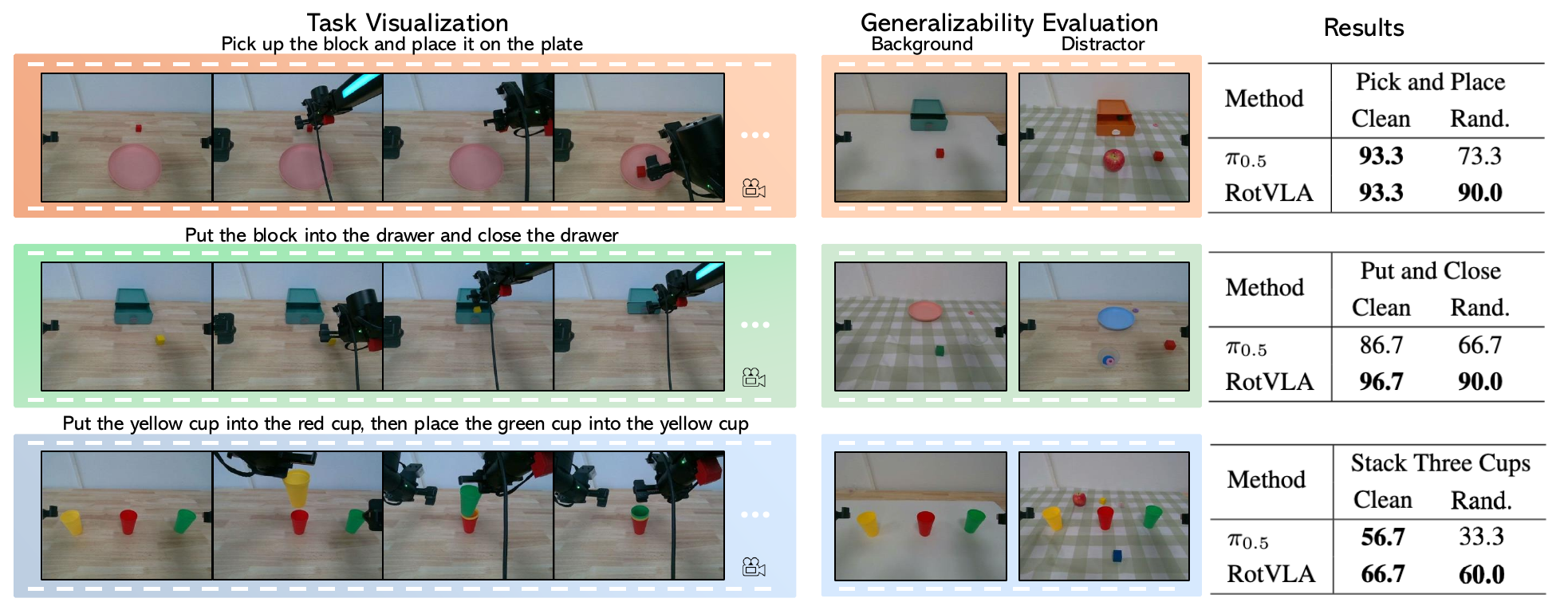}
  \caption{Visualization of real-world tasks and generalization setting.
  }
  \label{fig:real}
  \vspace{-10pt}
\end{figure}

The strong performance of RotVLA can be attributed to the VLA pretraining strategy guided by the continuous rotational latent action representation, which captures more representative smooth and compositional dynamics. The triplet learning framework further mitigates collapse of LAM by enforcing compositional consistency, thereby encouraging the learning of meaningful temporal representations. In addition, the latent actions function as high-level planners that guide action generation, leading to more stable long-horizon manipulation and improved multi-task generalization.

\subsection{Real World Experiment}
\label{sec: real world experiment}
We evaluate RotVLA on a dual-arm ARX R5 robotic platform across two single-arm tasks: (1) \textit{pick up the block and place it on the plate}, and (2) \textit{put the block into the drawer and close the drawer}, as well as one dual-arm task: (3) \textit{stack three cups}. We collect 100 demonstration trajectories for each task and train methods in a multi-task setting. We compare our method against $\pi_{0.5}$~\cite{intelligence2504pi0}. As shown in Fig.\ref{fig:real}, RotVLA consistently outperforms $\pi_{0.5}$~\cite{intelligence2504pi0} across all three tasks under standard evaluation settings. In the two single-arm tasks, RotVLA achieves \textbf{over 90\%} success rate. In the dual-arm cup stacking task, which requires bimanual motion and long-horizon planning, RotVLA achieves a substantially lower failure rate than the baseline. Under domain-shift conditions with varying backgrounds and the introduction of distractor objects, the performance gap becomes more pronounced. This robustness can be attributed to the continuous rotational latent action based pretraining and the latent planner finetuning, which captures high-level motion semantics rather than overfitting to background appearance. In terms of efficiency, on an NVIDIA H20 server, RotVLA achieves an inference latency of 79ms per step, compared to 61ms for $\pi_{0.5}$~\cite{intelligence2504pi0}, demonstrating competitive real-time performance.


\subsection{Analysis of Continuous Rotational Latent Action}

\begin{figure}[t]
  \centering
  \includegraphics[width=1\linewidth]{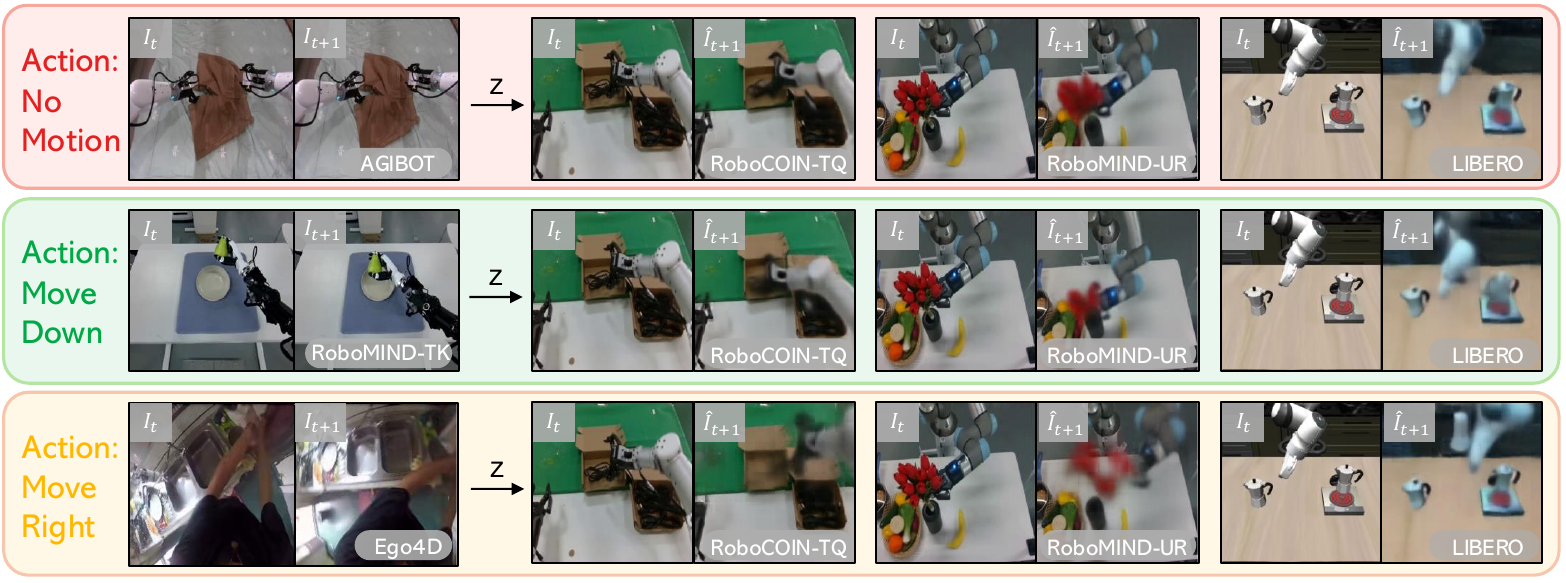}
  \caption{Illustration that the latent action extracted by one dataset can generalize to other seen and unseen datasets. RoboMIND-TK is short for Tien Kung robot in RoboMIND~\cite{wu2024robomind}. RoboCOIN-TQ is short for Tian Qin robot in RoboCOIN~\cite{wu2025robocoin}. 
  }
  \label{fig:lam_vis2}
\end{figure}

To demonstrate the effectiveness of the proposed LAM, we conduct cross-domain visualization experiments in Fig.\ref{fig:lam_vis2}. Specifically, we extract latent actions from one domain and apply them to different domains. For example, in the third row, a latent action extracted from the Ego4D human dataset, representing a right motion, generalizes well to the robotic platforms such as Tian Qin and UR5, and even to unseen datasets LIBERO~\cite{liu2023libero}, which were not used during training. Similarly, latent actions corresponding to motions such as moving right or remaining stationary exhibit strong cross-domain transferability. These results suggest that the learned latent actions capture high-level motion semantics that are independent of specific embodiments or visual appearances.


Beyond visualization results, we further analyze the proposed latent action from multiple perspectives.

\begin{figure}[t]
\centering

\begin{minipage}{0.44\textwidth}
\centering
\scriptsize
\renewcommand{\arraystretch}{1.4}
\begin{tabularx}{\linewidth}{l|>{\centering\arraybackslash}m{1.05cm}|>{\centering\arraybackslash}m{1.05cm}|>{\centering\arraybackslash}m{1.05cm}}
\hline
LAM Training & $\rm \hat{MSE}$  & $\rm \hat{MSE'}$ & $\Delta$ $\uparrow$ \\
\hline
Recon-Only~\cite{bu2025univla}  & 0.0029 &  0.0066 & 0.0037 \\
Triplet (Ours)  & 0.0030 &  0.0078 & \textbf{0.0048} \\
\hline
\end{tabularx}
\captionof{table}{Comparison of next-frame reconstruction and imagined-frame error. $\rm \hat{MSE}$ = $\rm MSE$ $(\hat{I}_{t+1}, I_{t+1})$, $\rm \hat{MSE'}$ = $\rm MSE$ $(\hat{I}'_{t+2}, I_{t+1})$.}
\label{table: degrade}
\end{minipage}
\hspace{0.0\textwidth}
\begin{minipage}{0.54\textwidth}
\centering
\includegraphics[height=2.7cm]{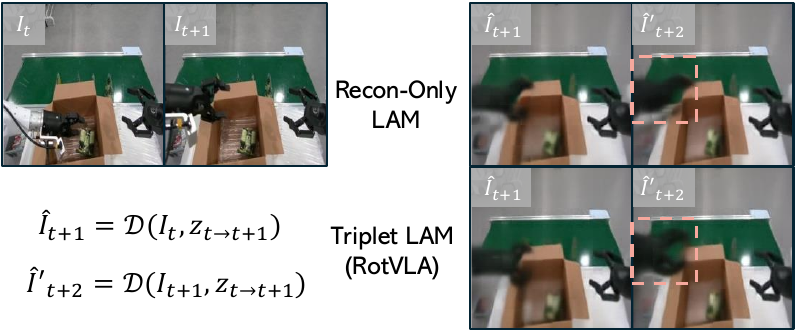}
\caption{Visualization results of $\hat{I}_{t+1}$ and $\hat{I}'_{t+2}$.}
\label{fig:degrade}
\end{minipage}

\end{figure}

\begin{table}[t]
\caption{Results of different latent action methods on LARY benchmark~\cite{nie2026lary}. RotVLA$^*$ denotes a variant whose LAM is pretrained on Open X-Embodiment~\cite{o2024open} and RoboMIND~\cite{wu2024robomind}. }
\centering
\small
\renewcommand{\arraystretch}{1.05}

\label{tab:lary-model-summary}
\setlength{\tabcolsep}{3.4pt}

\begin{tabular}{lcccccccc}
\toprule

\multirow{2}{*}{\textbf{Model}}  & \multicolumn{5}{c}{Regression MSE$\downarrow$} & \multicolumn{3}{c}{Classification Accuracy$\uparrow$}\\
\cmidrule(lr){2-6} \cmidrule(lr){7-9}
 & CALVIN& VLABench& RoboCOIN& AgiBot& Avg. & Robot & Human & Avg.\\
\midrule
LAPA~\cite{ye2024latent}   & 0.96& 0.95 &0.96 &1.00 & 0.97  &23.64 & 14.61&19.13\\
UniVLA~\cite{bu2025univla}  &    0.82 &0.74& 0.94 &0.97 & 0.87 &18.56 &  19.08 &18.82\\
villa-X~\cite{chen2025villa} & 0.86& 0.72 &0.94 &0.97 & 0.87 &29.90 & 17.80 &23.85\\
LAPA-DINOv3~\cite{nie2026lary} &    0.50 &0.25 &0.82 &0.84 & 0.63  &27.04 & 64.19&45.62\\

\midrule

RotVLA$^*$ &  0.38& 0.11 & 0.62& 0.68 & 0.27 & 61.26 & 58.13 & 59.70\\

RotVLA &   \textbf{0.30}& \textbf{0.08} & \textbf{0.43}& \textbf{0.44} & \textbf{0.20} & \textbf{67.62} & \textbf{74.33} & \textbf{70.98}\\
\bottomrule
\end{tabular}

\end{table}

\textbf{(1) Training stability.} {\textbf{Did the triplet learning framework mitigate the risk of degrading?}} To answer this question, we compare the proposed triplet learning framework with a conventional reconstruction-only baseline. Given two frames $I_t,I_{t+1}$ with interval $k$, we first extract the latent action, $z_{t\rightarrow t+1}=\mathcal{F}(I_t,I_{t+1})$ according to Eq.\ref{eq: latent action}. We then generate the reconstructed frame $\hat{I}_{t+1}=\mathcal{D}(I_t,z_{t\rightarrow t+1})$ and further apply the same latent action to $I_{t+1}$ to obtain an imagined frame $\hat{I}'_{t+2}=\mathcal{D}(I_{t+1},z_{t\rightarrow t+1})$. We compute the distances between $I_{t+1}$ and both $\hat{I}_{t+1}$ and $\hat{I}'_{t+2}$. Ideally, $\hat{I}_{t+1}$ should closely match the ground-truth $I_{t+1}$, while $\hat{I}'_{t+2}$ should reflect the application of the learned motion to $I_{t+1}$ and therefore differ from $I_{t+1}$. As shown in Table.\ref{table: degrade}, both methods achieve low reconstruction error for $I_{t+1}$. However, the reconstruction-only baseline produces significantly smaller differences between $I_{t+1}$ and $\hat{I}'_{t+2}$, indicating that the latent actions collapse to encoding frame appearance rather than true motion dynamics. Qualitative results in Fig.\ref{fig:degrade} further illustrate this effect. The LAM trained with the triplet framework successfully captures the dynamic of lifting the left arm and can extrapolate the next step from $I_{t+1}$. In contrast, the reconstruction-only model generates a frame nearly identical to $\hat{I}_{t+1}$, suggesting that its latent action encodes appearance cues instead of motion information.

\textbf{(2) Representation expressiveness.} \textbf{Is the proposed latent action more expressive than the previous methods?} To answer this question, we compare different latent action methods on the LARY benchmark~\cite{nie2026lary}, which evaluates latent action representations via linear probing. Results in Table~\ref{tab:lary-model-summary} show that RotVLA substantially outperforms existing methods, achieving both lower regression error for low-level robotic control and higher classification accuracy for high-level semantic actions. Moreover, we also train a variant, denoted RotVLA$^*$, whose LAM excludes datasets that overlap with the benchmark, including, AGIBOT-Beta, RoboCOIN, and Ego4D. As shown in Table~\ref{tab:lary-model-summary}, RotVLA$^*$ still significantly outperforms other LAMs on robotic data and achieves results comparable to LAPA-DINOv3 on human activity recognition, despite never being trained on any human video data. Furthermore, RotVLA consistently surpasses RotVLA$^*$ on held-out benchmarks such as CALVIN and VLABench, indicating that the LAM can effectively leverage additional diverse data to learn more transferable representations. These results demonstrate that the proposed latent action representation captures both fine-grained control dynamics and high-level semantic structure more effectively than prior approaches.

\begin{wrapfigure}{t}{0.5\textwidth}
\captionof{table}{Ablation results on different components.}
\centering
\small

\renewcommand{\arraystretch}{1.3}
\begin{tabularx}{\linewidth}{l|>{\centering\arraybackslash}m{0.58cm}>{\centering\arraybackslash}m{0.58cm}>{\centering\arraybackslash}m{0.58cm}>{\centering\arraybackslash}m{0.58cm}>{\centering\arraybackslash}m{0.58cm}}
\hline
Methods & Spatial & Object & Goal & Long & Avg. \\
\hline
w/o. Pretrain & 94.8 & 97.8& 94.2&  89.8 & 94.2 \\
w/o. Planner & 96.0 & \textbf{99.6} & 97.2 & 93.2 & 96.5\\
$n=8$ & 97.2 & 99.4 & 98.2 & 94.4 & 97.3\\
$n=32$  & \textbf{98.2} & \textbf{99.6} & 97.4 & 94.0 & 97.3\\
\hline 
Discrete & 95.4 & 99.4 &  96.6 &  89.6 & 95.3  \\
Cont.  & 96.6 & 99.4 & 97.2 & 93.6 & 96.7 \\
Cont. ${\rm SO}(n)$  & 97.0 & 99.2 & 97.6 & 94.2 & 97.0\\

RotVLA & \textbf{98.2} & \textbf{99.6} & \textbf{98.4} & \textbf{96.4} & \textbf{98.2} \\
\hline
\end{tabularx}

\label{table: ablation}
\end{wrapfigure}

\subsection{Ablation Study}
\label{sec: more ablation study}
In Table.\ref{table: ablation}, we conduct ablation studies to analyze key design choices in RotVLA, including the dimension of the latent action $n$, the triplet learning framework, the latent planner during finetuning.

\noindent \textbf{Latent planner.} We ablate the latent action supervision during finetuning. When latent action tokens are not preserved as a separate planning signal, performance drops, particularly on long-horizon tasks such as LIBERO-Long~\cite{liu2023libero}. 

\noindent \textbf{Dimension of the latent action.} We vary the $n$ across 8, 16 and 32. As shown in the last line, RotVLA with $n=16$ achieves the best performance. Smaller dimensions limit dynamic expressiveness, while larger dimensions increase optimization difficulty and increased computational cost.

\noindent \textbf{Latent Action Design.} We ablate different latent action designs. The discrete latent action paradigm~\cite{bu2025univla} brings only limited improvement over the w/o. Pretrain baseline due to the restricted representation capacity. Replacing it with continuous latent actions improves performance by 1.4\%. Adopting the ${\rm SO}(n)$ latent action alone yields marginal gains, suggesting that the structural constraint itself is insufficient. However, when the compositional supervision of $\hat{I}_{t+2}^{\rm comp}$ is further introduced on top of the ${\rm SO}(n)$ latent action, the performance improves by an additional 1.2\%. This result indicates that the triplet training framework encourages the LAM to capture motion dynamics rather than simply memorizing frame appearance.


\section{Conclusion}
We introduced RotVLA, a Vision-Language-Action framework built upon a continuous rotational latent action representation in ${\rm SO}(n)$. By replacing discrete latent tokens with structured rotations and introducing a triplet learning objective that enforces temporal compositionality, RotVLA learns meaningful and transferable motion dynamics while mitigating representation collapse. Built on a pretrained VLM with a flow-matching action expert, RotVLA treats latent actions as high-level planners that guide embodiment-specific control. Extensive experiments in simulation and real-world manipulation demonstrate strong generalization and competitive performance with 1.7B parameters. Our results suggest that incorporating geometric structure and continuity into latent action modeling is a promising direction toward scalable and generalizable robot foundation models. 

\begin{ack}
This paper makes use of Open X-Embodiment, AGIBOT-beta, RoboMIND, RoboCOIN, Ego4D, LIBERO and RoboTwin. The authors confirm that the use of the aforementioned datasets are solely for academic research purposes and not for any commercial activities.


\end{ack}

\clearpage

\bibliographystyle{unsrtnat}  
\bibliography{main}
\normalsize







\clearpage 

\appendix

\section{Limitation and Future Work}
\textbf{Parameter scalability.} RotVLA achieves state-of-the-art performance with 1.7B parameters, yet the overall model scale remains relatively moderate. In addition, the LAM model that provides supervision for the VLA model contains only 290M parameters, making it considerably smaller. Scaling both the LAM and VLA models may further improve performance.

\textbf{Learning from human videos.} Leveraging human videos has become a promising direction for addressing the scarcity of robotic data, and LAM provides an effective mechanism for generating supervision from large-scale unlabeled human videos. In RotVLA, human videos are used only as supplementary training data. Extending the framework to incorporate more human videos, or even pretraining solely on human videos, is an important direction for future work.

\section{Details of Continuous Latent Action}
We replace the VQ-VAE with SoftVQ~\cite{chen2025softvq} to preserve continuity in the latent space while retaining codebook structure. SoftVQ impose a soft categorical distribution on the posterior as:
\begin{equation}
  posterior:   q_\phi({\rm z|x}) = {\rm Softmax}(-\frac{\|z-\mathcal{C}\|_2}{\tau})
\end{equation}
\begin{equation}
    latent: {\rm z} = q_\phi({\rm z|x})\mathcal{C}
\end{equation}
where $\mathcal{C}$ represents the codebook with $K$ codewords, $\tau$ is the temperature parameter. When assuming the prior is a uniform distribution over the codebook:

\begin{align}
    \mathcal{L}_{\rm kl} &= \big(q_\phi({\rm z})\| p({\rm z})\big) \\
    & = \int q_\phi({\rm z})\big({\rm log} \ q_\phi({\rm z})-{\rm log} \ p({\rm z})\big) {\rm dz} \\
    & = H\big(q_\phi({\rm z})\big) - H\big(q_\phi({\rm z}),p({\rm z})\big),
\end{align}
where $p({\rm z})$ is the uniform piror that $p({\rm z}) \sim \mathcal{U}(0,K)$. In practice, the $\mathcal{L}_{\rm kl}$ is calculated as:
\begin{equation}
    \mathcal{L}_{\rm kl} = H\big(q_\phi({\rm z})\big) - H\big(\mathbb{E}_{{\rm x} \sim p({\rm x})} \ q_\phi({\rm z|x})\big),
\end{equation}
which represents the $\mathcal{L}_{\rm soft}$ in Eq.13 in the main paper.

\begin{table}[h]
\centering
\small
\caption{Detailed hyperparameters of RotVLA}
\label{tab: hyperparameter}
\renewcommand{\arraystretch}{1.3}
\begin{tabular}{lccc}
\hline
\multirow{2}{*}{\textbf{Hyperparameters} } & \multicolumn{3}{c}{RotVLA} \\
\cline{2-4}
 & \textbf{Stage \text{I}} & \textbf{Stage \text{II}} & \textbf{Stage \text{III}} \\
\hline
Learning rate      & $1.0 \times 10^{-4}$ &$1.0 \times 10^{-4}$ & $1.0 \times 10^{-4}$  \\
LR scheduler       & Constant               & Cosine              & Cosine \\
Weight decay       & 0.01                  & 0.001               & 0.001 \\
Gradient clip      & -                  & 1.0                 & 1.0 \\
Optimizer          & \multicolumn{3}{c}{AdamW($\beta_1=0.9, \beta_2=0.999$)} \\
Warm-up steps      & 0                  & 5k              & 5k \\
Training steps     & 200k                & 200k             & 80-120k \\
Batch size         & 256                  & 256              & 128 \\
Image Size    & 224                  & 224              & 224 \\
Data Augumentation & - & \multicolumn{2}{c}{ColorJitter}   \\
Trainable Parameters & 0.20B                 & 1.36B              & 1.36B \\

Total Parameters & 0.29B                 & 1.65B              & 1.65B \\
\hline
\end{tabular}
\end{table}

\section{Implementation Details}
During stage \text{I} and \text{II}, we pretrain both LAM and RotVLA for 200k steps, with the batchsize of 256. During downstream finetuning, RotVLA is trained with a batch size of 128. The training steps for LIBERO, RoboTwin2.0 and real-world experiment is 80k, 120k and 40k respectively. The latent action model is frozen after stage \text{I}. Detailed hyperparameters are list in Table.\ref{tab: hyperparameter}.

\section{Real-World Experiments}
\begin{figure}[t]
  \centering
  \includegraphics[width=1\linewidth]{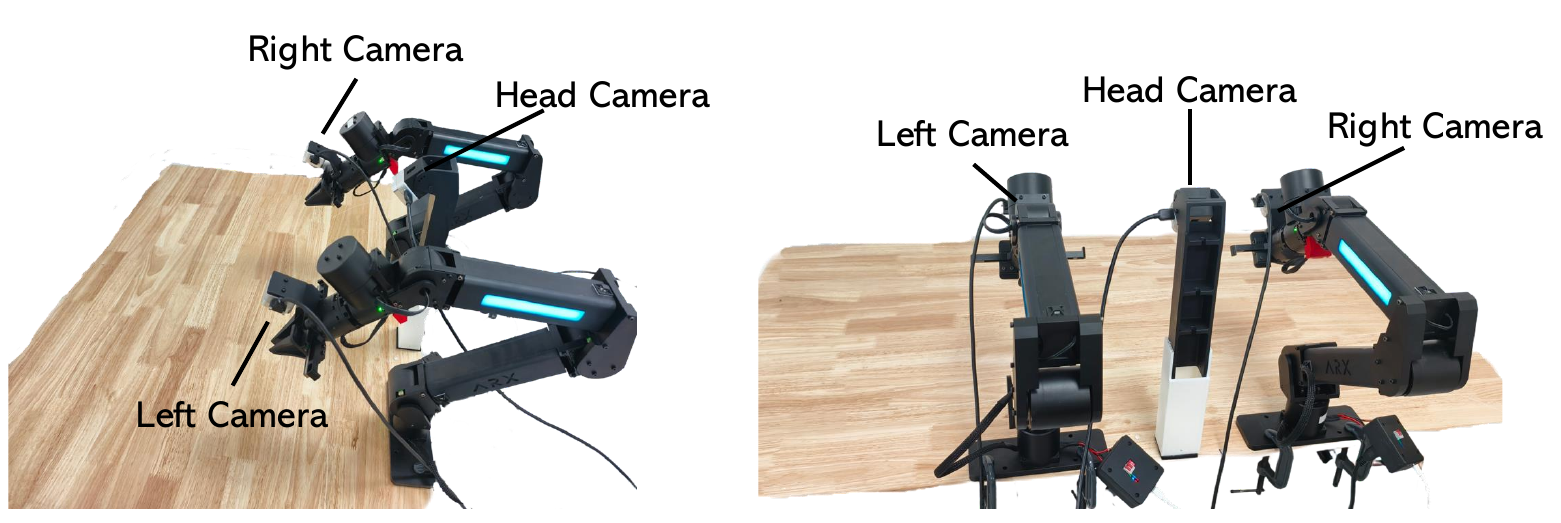}
  \caption{The robotic platform used in real-world experiments.}
  \label{fig:robot platform}
\end{figure}

Real-world experiments are conducted on a dual-arm ARX R5 robotic platform equipped with three RealSense D405 cameras: one mounted on the head and two on the wrists. The detailed hardware setup is shown in Fig.\ref{fig:robot platform}.  We collect 30 minutes of teleoperation data per task at 30Hz. A single model is trained across all three tasks, conditioned on different language instructions.

We evaluate RotVLA on three tasks: (1) Pick up the block and place it on the plate. (2) Put the block into the drawer and close the drawer. (3) Put the yellow cup into the red cup, then place the green cup into the yellow cup. Qualitative results of RotVLA are provided in the \texttt{videos/} folder of the supplementary material. Each tasks is evaluated for 30 trials each with clean and randomized settings. For the randomized setting: 10 trials with background changes (two out-of-distribution backgrounds: plain white and patterned), 10 trials with distractor objects, and 10 trials with both background changes and distractors.

\section{More Visualization of Continuous Rotational Latent Action}
\begin{figure}[t]
  \centering
  \includegraphics[width=1\linewidth]{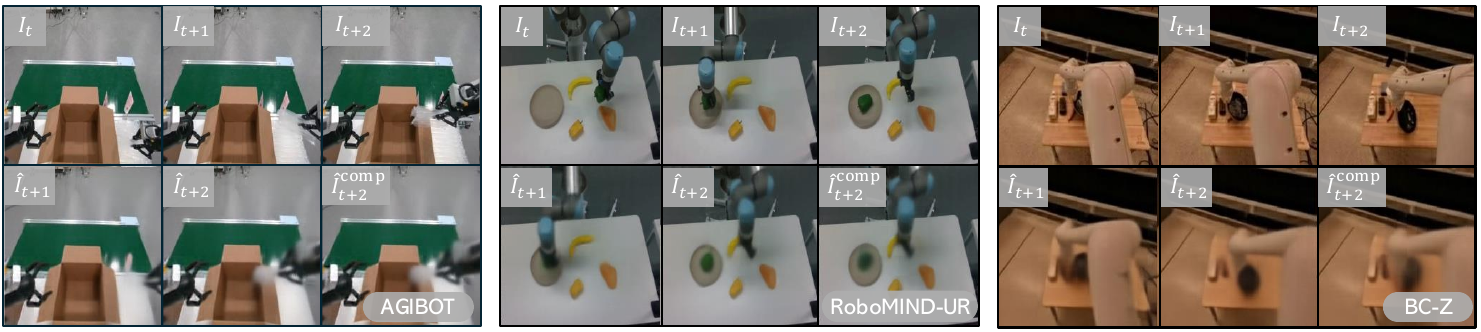}
  \caption{Illustration of reconstruction of $\hat{I}_{t+1},\hat{I}_{t+2}$ and also the composed $\hat{I}_{t+2}^{\rm comp}$ on AGIBOT~\cite{bu2025agibot}, RoboMIND~\cite{wu2024robomind} and BC-Z~\cite{jang2022bc} datasets.
  }
  \label{fig:lam_vis1}
\end{figure}

To demonstrate the effectiveness of the proposed LAM, we visualize reconstruction results in Fig.\ref{fig:lam_vis1}. The first line shows the sampled frame $I_t,I_{t+1}$ and $I_{t+2}$, while the second line presents the corresponding generated frames $\hat{I}_{t+1},\hat{I}_{t+2}$, and the composed prediction $\hat{I}_{t+2}^{\rm comp}$, obtained by applying the composed latent action to $I_t$. The results indicate that the LAM effectively captures the dynamic evolution of the robot across frames. Importantly, the composed latent action successfully represents the two-step transition from $I_t$ to $I_{t+2}$, demonstrating compositional consistency in the learned latent space.

\begin{figure}[t]
  \centering
  \includegraphics[width=0.6\linewidth]{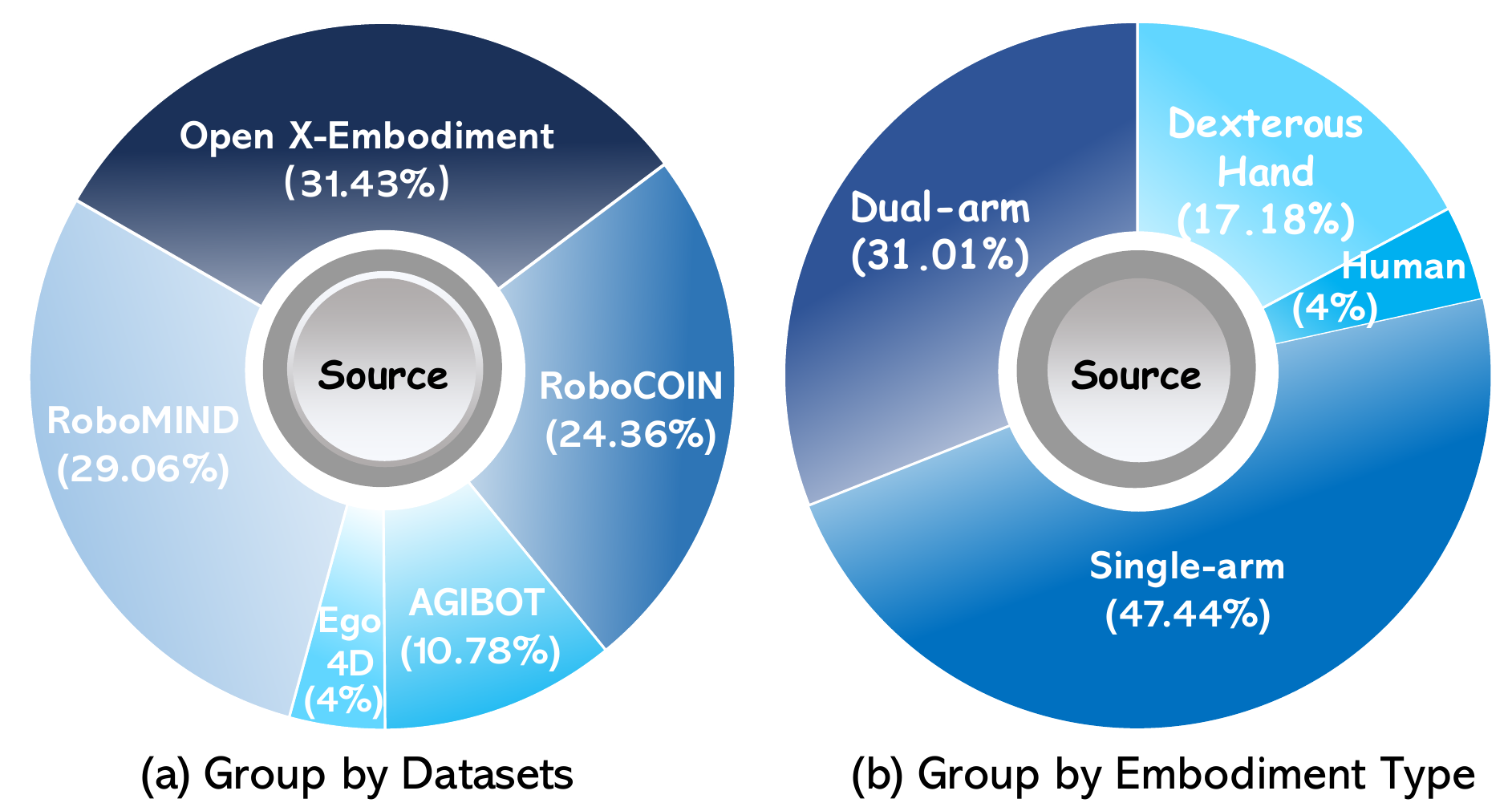}
  \caption{Statistics of the pretraining data used by RotVLA, grouped by dataset and embodiment type.
  }
  \label{fig:data source}
\end{figure}

\section{Detailed Results on RoboTwin2.0}
In Table.\ref{tab:robottwin2_sim_rotvla} we provide detailed results of RoboTwin2.0 benchmark~\cite{chen2025robotwin} across 50 tasks. RotVLA achieves the success rate of \textbf{89.6\%} / \textbf{88.5\%} on the clean and randomized settings.

\section{Data Statistic}
We curate a training dataset totaling over 1700 hours from Open X-Embodiment~\cite{o2024open}, AGIBOT~\cite{bu2025agibot}, RoboCOIN~\cite{wu2025robocoin}, RoboMIND~\cite{wu2024robomind}, and Ego4D~\cite{grauman2022ego4d}, consisting of dual-arm and single-arm robot data, dexterous hands data as well as egocentric human videos. Detailed statistics of the different data sources are listed in Table.\ref{tab:dataset norm final}. When grouped by dataset source (Fig.\ref{fig:data source}(a)), the proportions are: Open X-Embodiment (31.43\%), RoboMIND (29.06\%), RoboCOIN (24.36\%), AGIBOT (10.78\%), and Ego4D (4.37\%). When grouped by embodiment type (Fig.\ref{fig:data source}(b)), the proportions are: dexterous hands (17.18\%), human videos (4.37\%), single-arm robots (47.44\%), and dual-arm robots (31.01\%).

\begin{figure}[t]
  \centering
  \includegraphics[width=0.6\linewidth]{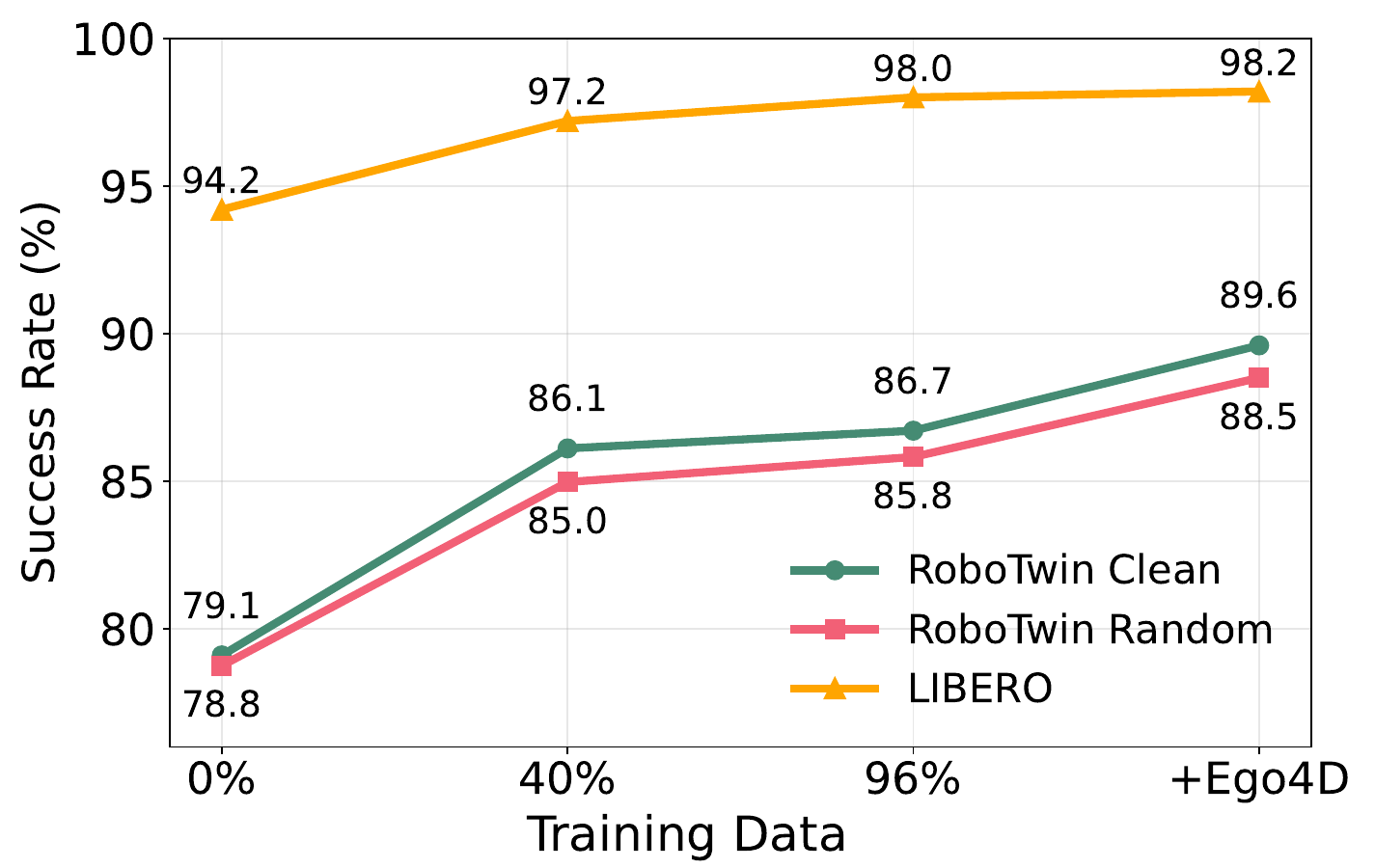}
  \caption{Impact of pretraining data scale on performance.
  }
  \label{fig:data scale}
\end{figure}

\section{Data scalability}
We analyze how RotVLA benefits from increasing the scale of pretraining data. When pretrained on only 40\% of the robot data, RotVLA achieves a 3\% improvement on LIBERO and over 7\% on RoboTwin compared to training without pretraining. Expanding to the full robot datasets, including Open X-Embodiment~\cite{o2024open}, AGIBOT~\cite{bu2025agibot}, RoboMIND~\cite{wu2024robomind}, and RoboCOIN~\cite{wu2025robocoin}, further improves performance. Moreover, incorporating human video dataset, Ego4D~\cite{grauman2022ego4d}, provides additional gains, particularly on RoboTwin2.0. These results demonstrate that scaling both the quantity and diversity of pretraining data consistently enhances performance, highlighting strong data scalability.

\begin{table*}[h]
\caption{Evaluation on RoboTwin 2.0 simulation benchmark.}
\centering
\renewcommand{\arraystretch}{1.3}
\resizebox{0.85\textwidth}{!}{%
\begin{tabular}{lcc|lcc}
\hline
\textbf{Task} & \textbf{Clean} & \textbf{Rand.} & \textbf{Task} & \textbf{Clean} & \textbf{Rand.} \\
\hline
\textit{Adjust Bottle} & 100\% & 100\% & \textit{Place Can Basket} & 83\% & 76\% \\
\textit{Beat Block Hammer} & 92\% & 85\% & \textit{Place Cans Plasticbox} & 100\% & 100\% \\
\textit{Blocks Ranking Rgb} & 98\% & 96\% & \textit{Place Container Plate} & 99\% & 100\% \\
\textit{Blocks Ranking Size} & 78\% & 77\% & \textit{Place Dual Shoes} & 93\% & 96\% \\
\textit{Click Alarmclock} & 47\% & 46\% & \textit{Place Empty Cup} & 100\% & 99\% \\
\textit{Click Bell} & 97\% & 92\% & \textit{Place Fan} & 97\% & 97\% \\
\textit{Dump Bin Bigbin} & 97\% & 98\% & \textit{Place Mouse Pad} & 83\% & 87\% \\
\textit{Grab Roller} & 100\% & 100\% & \textit{Place Object Basket} & 91\% & 90\% \\
\textit{Handover Block} & 95\% & 84\% & \textit{Place Object Scale} & 92\% & 93\% \\
\textit{Handover Mic} & 100\% & 99\% & \textit{Place Object Stand} & 97\% & 100\% \\
\textit{Hanging Mug} & 44\% & 52\% & \textit{Place Phone Stand} & 94\% & 94\% \\
\textit{Lift Pot} & 98\% & 100\% & \textit{Place Shoe} & 99\% & 100\% \\
\textit{Move Can Pot} & 95\% & 95\% & \textit{Press Stapler} & 83\% & 81\% \\
\textit{Move Pillbottle Pad} & 95\% & 99\% & \textit{Put Bottles Dustbin} & 91\% & 88\% \\
\textit{Move Playingcard Away} & 99\% & 99\% & \textit{Put Object Cabinet} & 87\% & 82\% \\
\textit{Move Stapler Pad} & 86\% & 84\% & \textit{Rotate Qrcode} & 95\% & 95\% \\
\textit{Open Laptop} & 98\% & 99\% & \textit{Scan Object} & 88\% & 85\% \\
\textit{Open Microwave} & 36\% & 12\% & \textit{Shake Bottle Horizontally} & 99\% & 99\% \\
\textit{Pick Diverse Bottles} & 88\% & 88\% & \textit{Shake Bottle} & 100\% & 100\% \\
\textit{Pick Dual Bottles} & 89\% & 88\% & \textit{Stack Blocks Three} & 90\% & 86\% \\
\textit{Place A2b Left} & 93\% & 95\% & \textit{Stack Blocks Two} & 99\% & 99\% \\
\textit{Place A2b Right} & 97\% & 96\% & \textit{Stack Bowls Three} & 92\% & 88\% \\
\textit{Place Bread Basket} & 96\% & 93\% & \textit{Stack Bowls Two} & 97\% & 96\% \\
\textit{Place Bread Skillet} & 89\% & 88\% & \textit{Stamp Seal} & 81\% & 83\% \\
\textit{Place Burger Fries} & 99\% & 98\% & \textit{Turn Switch} & 46\% & 46\% \\
\hline
{\textbf{Average}}&  & \multicolumn{2}{c}{}& \textbf{89.6\%} & \textbf{88.5\%} \\
\hline
\end{tabular}%
}

\label{tab:robottwin2_sim_rotvla}
\end{table*}

\clearpage

\captionsetup{type=table,skip=1pt}
\captionof{table}{The data mixture of RotVLA.}
\label{tab:dataset norm final}
\renewcommand{\arraystretch}{1.35}
\setlength{\tabcolsep}{6pt}
\small
\begin{longtable}{l|>{\raggedright\arraybackslash}p{3.5cm}|r|l}
\hline
\textbf{Source} & \textbf{Robot / Dataset}  & \textbf{Norm} & \textbf{Type} \\
\hline

\multirow{1}{*}{\textbf{Ego4D}~\cite{grauman2022ego4d}} & Human & 4.37\% & Human \\
\hline

\multirow{9}{*}{\textbf{RoboCOIN}~\cite{wu2025robocoin}} & RMC-AIDA-L~\cite{realman_robotics} &  3.08\% & Dual-Arm \\
 & Agilex Split ALOHA~\cite{agilex_cobot_magic} & 1.94\% & Dual-Arm \\
  & Agilex Cobot Magic~\cite{agilex_cobot_magic} &  4.65\% & Dual-Arm \\
 & Galaxea R1 Lite~\cite{galaxea_r1_lite} & 3.75\% & Dual-Arm \\

 & AgiBot G1~\cite{agibot_g1} &  2.13\% & Dual-Arm \\
 & Airbot MMK2~\cite{airbot_mmk2} &  3.83\% & Dexterous Hand \\
 & Unitree G1edu-u3~\cite{unitree_g1_2024} &  2.55\% & Dexterous Hand \\
 & Tianqing A2~\cite{tqartisan_a2} & 2.43\% & Dual-Arm \\
\hline

\multirow{4}{*}{\textbf{RoboMIND}~\cite{wu2024robomind}} & UR5e~\cite{ur5e2024}  & 8.22\% & Single-Arm \\
 & Franka Emika Panda~\cite{franka_panda} & 7.79\% & Single-Arm \\
 & Tien Kung~\cite{xhumanoid2024}  & 2.26\% & Dual-Arm \\
 & Tien Kung~\cite{xhumanoid2024}  & 10.79\% & Dexterous Hand \\
\hline

\multirow{25}{*}{\shortstack{\textbf{Open X-} \\ \textbf{Embodiment}~\cite{o2024open}}} & BC-Z~\cite{jang2022bc} & 3.42\% & Single-Arm \\
 & Furniture Bench~\cite{heo2025furniturebench} &  2.49\% & Single-Arm \\
 & Fractal~\cite{brohan2022rt} & 5.92\% & Single-Arm \\
 & Kuka~\cite{kalashnikov2018scalable} & 1.16\% & Single-Arm \\
 & DobbE~\cite{shafiullah2023bringing} & 0.71\% & Single-Arm \\
 & FMB Datasets~\cite{luo2025fmb} & 3.60\% & Single-Arm \\
 & Bridge~\cite{walke2023bridgedata} & 2.57\% & Single-Arm \\
 & UTAustin Mutex~\cite{shah2023mutex} & 1.14\% & Single-Arm \\
 & Stanford Hydra Dataset~\cite{belkhale2023hydra} & 2.27\% & Single-Arm \\
 & Austin Sailor Dataset~\cite{nasiriany2022learning} &  1.12\% & Single-Arm \\
& Austin Sirius Dataset~\cite{liu2025robot}&  0.89\% & Single-Arm \\
 & Toto~\cite{zhou2023train} & 0.93\% & Single-Arm \\   
 & Taco Play~\cite{mees2022grounding} &  1.35\% & Single-Arm \\
 & Roboturk~\cite{mandlekar2018roboturk} &  1.07\% & Single-Arm \\
 & CMU Franka Pick-Insert~\cite{saxena2023multi} & 0.46\% & Single-Arm \\
 & Berkeley Autolab UR5~\cite{BerkeleyUR5Website} & 0.56\% & Single-Arm \\
 & Jaco Play~\cite{dass2023jacoplay} &  0.22\% & Single-Arm \\
 & Viola~\cite{zhu2023viola} &  0.44\% & Single-Arm \\
 & Berkeley Fanuc Manipulation~\cite{zhu2023fanuc} & 0.40\% & Single-Arm \\
 & Berkeley Cable Routing~\cite{luo2024multistage} &  0.12\% & Single-Arm \\
 & NYU Franka Play~\cite{cui2022play}&  0.33\% & Single-Arm \\
 & Austin Bud~\cite{zhu2022bottom} & 0.11\% & Single-Arm \\
 & CMU Stretch~\cite{mendonca2023structured} & 0.08\% & Single-Arm \\
 & DLR EDAN Shared Control~\cite{quere2020shared} &0.03\% & Single-Arm \\
 & UCSD Kitchen~\cite{ucsd_kitchens} & 0.03\% & Single-Arm \\
\hline

 \textbf{AGIBOT}~\cite{bu2025agibot} & AgiBot G1~\cite{agibot_g1} & 10.78\% & Dual-Arm \\

\hline

\multicolumn{2}{l|}{\textbf{Total}} & \textbf{100.00\%} & \\
\hline
\end{longtable}

\end{document}